\documentclass{article}

\usepackage[preprint]{neurips_2022}

\usepackage[utf8]{inputenc} 
\usepackage[T1]{fontenc}    
\usepackage{hyperref}       
\usepackage{url}            
\usepackage{booktabs}       
\usepackage{amsfonts}       
\usepackage{nicefrac}       
\usepackage{microtype}      
\usepackage{xcolor}         
\usepackage{graphicx}   
\usepackage{amsmath}
\usepackage{subcaption}
\usepackage{float}
\usepackage{subcaption} 
\usepackage{mlmath} 
\usepackage{ulem}
\usepackage{authblk}
\usepackage{authblk}

\newcommand*{\email}[1]{\texttt{#1}}

\DeclareMathOperator*{\argmax}{arg\,max}

\title{An Experimental Comparison Between Temporal Difference and Residual Gradient with Neural Network Approximation}

\author{
Shuyu Yin\textsuperscript{\rm 1}, 
Tao Luo\textsuperscript{\rm 2,3}, 
Peilin Liu\textsuperscript{\rm 1}, 
Zhi-Qin John Xu\textsuperscript{\rm 2}\thanks{Corresponding author: xuzhiqin@sjtu.edu.cn.}, 
\\
\textsuperscript{\rm 1} School of Electronic Information and Electrical Engineering, Shanghai Jiao Tong University \\
\textsuperscript{\rm 2} School of Mathematical Sciences, Institute of Natural Sciences, MOE-LSC \\ 
and Qing Yuan Research Institute, Shanghai Jiao Tong University \\
\textsuperscript{\rm 3} CMA-Shanghai\\
\email{\{shuyu.yin, luotao41, liupeilin, xuzhiqin\}@sjtu.edu.cn}
}

\begin{document}

\maketitle

\begin{abstract} \label{abstraction}
Gradient descent or its variants are popular in training  neural networks. However, in deep Q-learning with neural network approximation, a type of reinforcement learning, gradient descent (also known as Residual Gradient (RG)) is barely used to solve Bellman residual minimization problem. On the contrary, Temporal Difference (TD), an incomplete gradient descent method prevails. In this work, we perform extensive experiments to show that TD outperforms RG, that is, when the training leads to a small Bellman residual error, the solution found by TD has a better policy and is more robust against the perturbation of neural network parameters. We further use experiments to reveal a key difference between reinforcement learning and supervised learning, that is, a small Bellman residual error can correspond to a bad policy in reinforcement learning while the test loss function in supervised learning is a standard index to indicate the performance. We also empirically examine that the missing term in TD is a key reason why RG performs badly. Our work shows that the performance of a deep Q-learning solution is closely related to the training dynamics and how an incomplete gradient descent method can find a good policy is interesting for future study.
\end{abstract}

\section{Introduction} \label{introduction}
In recent years, many Deep Reinforcement Learning (DRL) applications appear in gaming \citep{mnih2013playing,silver2017mastering,vinyals2019grandmaster}, recommendation system (\cite{deng2021unified}), combinatorial optimization (\cite{bello2016neural,khalil2017learning}), etc. These successful applications reveal the huge potential of DRL. Temporal Difference (TD, \cite{sutton1988learning}) method is an important component for constructing DRL algorithms, which mainly appears in deep Q-learning-like algorithms, for example
\cite{mnih2013playing,van2016deep,wang2016dueling,hessel2018rainbow}
deal with problems with discrete action space. TD method is extended to continuous action space by using actor-critic framework (\cite{gu2016continuous}). In deep Q-learning, the state-action value function $Q$ is parameterized by a hypothesis function, e.g., neural network. To optimize the parameters, the loss function is defined by the mean squared error between the prediction $Q$ and its true value. However, the true value is untraceable during the training, which is often estimated by Monte-Carlo method or bootstrapping method based on the Bellman equation. When we use a bootstrapping method to estimate the true value, there are two different optimization methods emerge. In the first one, we plugin the estimation of the true value to the loss function. In this case, the estimation also contains the state-action value function of some states, therefore, it also depends on the parameters. We then perform gradient descent on the loss function, where the gradient is performed on the prediction and the estimation. This is an exact gradient descent method, known as Residual Gradient (RG) method (\cite{baird1995residual,duan2021risk}). 
In the second one, also known as the TD method, we plugin the estimation of the true value after we  perform the gradient descent on the loss function. That is, the gradient is not operated on the true value but only on the prediction one. 
An interesting question arises that why the gradient descent, so popular in deep learning, is not commonly used in deep Q-learning, while the TD method, an incomplete gradient descent method, prevails. A more important issue is what special characteristics of deep Q-learning, different from training a neural network in supervised learning, make TD popular in deep Q-learning? 


In this work, we are trying to answer the above two questions. 
We conduct experiments in four different environments and in order to walk around double sampling issue we only consider deterministic environments, including Grid World, Cart Pole, Mountain Car, and Acrobot. We use a neural network to parameterize the state-action value function. 
Also, we follow the off-policy scheme, which uses a fixed data set to train the neural network, the data sets are sampled before training. In order to further reduce randomness, we use the whole set of data to do gradient descent. Our contribution is summarized as follows. 
\begin{itemize}
    \item TD can learn better policy compared to the RG method in our setting. The goodness is twofold: first, the policy learned by TD has a larger accumulated reward. Second, the neural network is more robust and trained by TD than RG when introducing perturbation on parameters. 
    \item Small Bellman residual error does not indicate a good policy, state values also need to be considered. In the simple problems we considered, policies can be divided into good and bad regimes according to state value when Bellman residual error is small. Both TD and RG can achieve small losses, however, we empirically find that the RG method is more likely to learn a policy in a bad regime.
    \item The missing term of TD is a key reason why RG performs badly. We empirically demonstrated that the missing term of TD, which is defined as backward-semi gradient, leads the RG to go to a bad policy regime.
\end{itemize}

The first contribution points out that TD is better than RG, which answers why TD is more popular than RG. The second contribution shows a special characteristic of deep Q-learning, i.e., a small loss does not indicate good performance, which is an answer to the second question. The third contribution further analyzes the special characteristics of RG. Our work makes a step towards a better understanding of the learning process of deep Q-learning.


\section{Related works}

We reviewed several basic empirical comparisons and theoretical analyzes between TD and RG.
The RG method was officially proposed by \cite{baird1995residual}. It is an alternative approach to solving Bellman residual minimization problem. However,  \cite{baird1995residual} does not give enough comparison of the performance between RG and TD.
\cite{schoknecht2003td} uses spectrum radius analysis to prove that TD converges faster than RG under synchronous, linear function approximation and one-hot embedding settings. Because the proof depends on the iterative structure of the linear model and synchronous structure, it is hard to extend their argument into the deep Q-learning setting.
\cite{li2008worst} uses techniques in stochastic approximation and prove that given $N$ training samples, if the model is trained for $N$ steps, RG would have a smaller total Bellman residual error, which is defined as the summation of the Bellman residual error with single training data in each step. This theorem is proved under the asynchronous setting and the iterative structure of linear approximation, which is also hard to extend to deep Q-learning problems.
\cite{scherrer2010should} constructs relationships between the loss function of RG and TD and the learned solution of RG and TD in value iteration and linear approximation setting. The result heavily relays on the convexity of the loss function and the simple geometric meaning of function space of linear function. So all the conclusions it \cite{scherrer2010should} cannot be applied to a deep Q-learning setting.
All of the theoretical analysis mentioned above not only depends on the linear structure but also equate small loss with good performance. However, this work suggests small loss doesn't indicate good policy.
\cite{zhang2019deep} combine RG and Deep Deterministic Policy Gradient (DDPG) (\cite{lillicrap2015continuous}) to construct Bi-Res-DDPG method. This combination can improve the performance in some Gym and Atari environments. With modification on RG, it is likely to improve RG in training RL models.

To our best knowledge, the only work that compares TD and RG in the deep Q-learning setting is \cite{saleh2019deterministic}. The main difference between these two papers is: this work provide a consistent explanation of the performance difference of RG and TD in both on-policy and off-policy setting using good and bad policy regime, however, \cite{saleh2019deterministic} only explain the performance difference in off-policy setting. Besides, the example provided in \cite{saleh2019deterministic} Figure.4 is constructed using linear approximation and uses an identity matrix as a feature matrix, which is actually a tabular setting. So this example cannot provide enough evidence that "distribution mismatch" heavily affects the performance of RG in the DQN setting. In addition, this work shows that the learning dynamic of "backward-semi gradient" is a key factor to explain the bad performance of RG, which is also not mentioned in \cite{saleh2019deterministic}.

\section{Preliminaries}

\subsection{Basic concepts}

Markov Decision Process is defined as $\mathcal{M}(S, A, f, r, \gamma)$, where $S$ is the state space, $s \in S$ is a state, $A$ is the action space and it is a finite set, $a\in A$ is an action, $f: S \times A \to S$ is the state transition function, $r: S \times A \to \mathbb{R}$ is the reward function and $\gamma \in [0, 1)$ is the discount factor. Reward function can be divided into two types: step reward $r_s(s,a)$ represents the reward between two non-terminal states and terminal reward $r_T(s,a)$ represents the reward between non-terminal state and terminal state. We defined the state-action value function as $Q: S \times A \to \mathbb{R}$ and state value function $V(\cdot) = \max_{a \in A}Q(\cdot,a)$.
Given training data set $D_N = \{ z_i \}_{i=1}^{N} = \{ (s_i, a_i, s'_i, r_i) \}_{i=1}^{N}$ where $s_i, s'_i \in S, a_i \in A, f(s_i,a_i)=s'_i$ and $r_i=r(s_i,a_i)$, and use $\vtheta$ to parameterize state-action value function. The empirical Bellman residual loss function is defined as
\begin{align}
    \mathcal{L} = \frac{1}{N} \sum_{i=1}^N(Q_{\vtheta}(s_i,a_i) - r(s_i, a_i) -\gamma \max_{a' \in A} Q_{\vtheta}(s'_i, a'))^2.
\end{align}

The gradient of this loss function with respect to parameter $\vtheta$ is 
\begin{align}
    \nabla \mathcal{L}_{\text{true}} = \frac{2}{N} \sum_{i=1}^N (Q_{\vtheta}(s_i,a_i) - r(s_i, a_i) -\gamma \max_{a' \in A} Q_{\vtheta}(s'_i, a'))(\nabla Q_{\vtheta}(s_i,a_i) - \gamma \nabla \max_{a' \in A} Q_{\vtheta}(s'_i, a')),
\end{align}
where $\nabla \max_{a' \in A} Q(s',a')$ is defined as $\nabla Q(s',a')$ such that $a' = \argmax_{a''} Q(s',a'')$.
This gradient uses the exact gradient form for the RG method, so-called true gradient. $\eta$ is the learning rate. On the other hand, for the TD method, the semi gradient is defined as
\begin{align}
    \nabla \mathcal{L}_{\text{semi}} = \frac{2}{N} \sum_{i=1}^N (Q_{\vtheta}(s_i,a_i) - r(s_i, a_i) -\gamma \max_{a' \in A} Q_{\vtheta}(s'_i, a'))\nabla Q_{\vtheta}(s_i,a_i),
\end{align}
We did not consider the loss function that cooperates with the target network in this work. Because when using target network, the RG method cannot be defined. Besides, the motivation to add the target network is to stable the training behavior of TD, however, we do not consider the hyper-parameter that makes TD unstable.

\subsection{Experiment setting}

\begin{figure}[htb]
\centering
\includegraphics[width=0.3\linewidth]{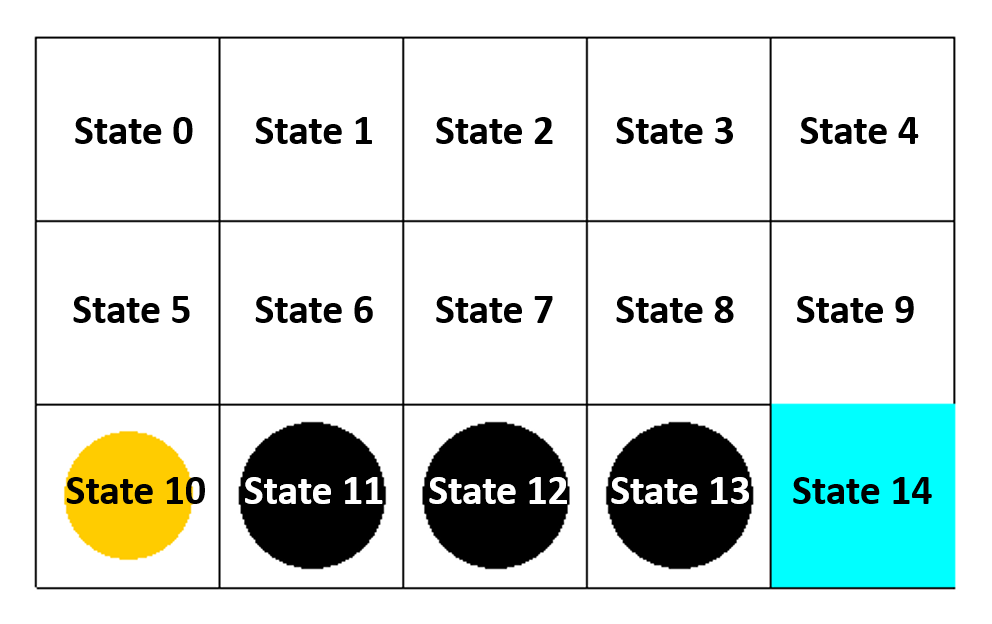}
\caption{The geometric setting of the Grid World environment. The yellow circle represents the starting point, the black circles represent traps, and the blue rectangle represents the goal. }
\label{fig: grid world env}
\end{figure}

We conduct experiments on the deterministic environments including discrete state and continuous state settings. There are four environments in total: one grid world environment and three gym classic control environments (\cite{brockman2016openai}): Mountain Car, Cart Pole, and Acrobot. The project of these environments is under MIT license, we can use it for research purpose. We read these codes through, there are no personal information or offensive content in these codes.\label{asset citation} The geometric structure of the Grid World environment is shown in Figure \ref{fig: grid world env}. Traps and the goal are terminal states. The action set is \{Up, Down, Left, Right\}. Only reaching the goal can bring a positive reward, step into a trap leading to a penalty and each step has a small penalty. If the agent hit the boundary, e.g., the agent goes up when it is in the first row, it will stay at its current state. The number represents the state index.

For the Grid World environment, discount factor $\gamma = 0.95$, the terminal reward for the goal state is +1, the terminal reward for trap states is -1, step reward is -0.01. For the Mountain Car environment, discount factor $\gamma = 0.93$, terminal reward is +1, step reward is -1. For the Cart Pole environment, discount factor $\gamma = 0.87$, terminal reward is -5, step reward is +1. For the Acrobot environment, discount factor $\gamma=0.98$, the terminal reward is 0, step reward is -1. In order to keep the training stability of the TD method, the discount factors can be randomly chosen as long as they are not very close to one to ensure both training algorithms converge. 

To control the randomness in exploration, we trained the model with fix data set under the off-policy scheme. For the Grid World environment, the data set is generated by the Cartesian product of non-terminal states and all actions. For the other three Gym environments, the data set is sampled with a random seed, the random seed is used to choose random actions and control the initial state of the environment at the beginning of the sampling process. For the Cart Pole environment, we collect samples from the first 100 epochs of random play with random seed 100, the total size of the data set is 2184. For the Mountain Car environment, we collect samples from the first 3 epochs of random play with random seed 550, the total number of data set is 14826. For the Acrobot environment, we collect samples from the first 10 epochs of random play with random seed 190, the total number of data set is 13771. We use the whole set of data to do gradient descent.

\label{detailed training parameters}
Also, we only change the parameter initialization in different trials and keep other hyper-parameters the same, like learning rate, training epoch, and optimization algorithms. In each trial, we use both TD and RG to train the neural network models with the same initial parameter values and other training hyper-parameters. All experiments use Adam as an optimizer. For the Grid World environment, the training step is 30000 and for the other environments, the training step is 50000. The initial learning rate of the Grid World environment is $10^{-4}$ and $10^{-5}$ for the other environments. The learning rate decays to its $85\%$ and $75\%$ for each $3000$ steps for the Grid World environment and the other environments, respectively. We use an NVIDIA GeForce RTX 3080 GPU to train the model. We use a four-layer fully connected neural network to parameterize the state-action value function $Q$, the hidden unit number in each layer is 512, 1024, 1024, and use ReLU as an activation function. \label{training resources}

\label{our limitaition}
In addition, for the pre-terminal state, which is the state before the terminal state, we do not force it by a fixed value during the training. We limit our discussion around this setting in this work, however, the methodology of analyzing the regime of policy can be applied to wider settings. 



\section{TD performs better compared to RG}

\subsection{TD learns a better policy} \label{sec:TDgoodexp}

\begin{figure}[htb]
\centering
\begin{subfigure}{.24\textwidth}
  \centering
  \includegraphics[width=1\linewidth]{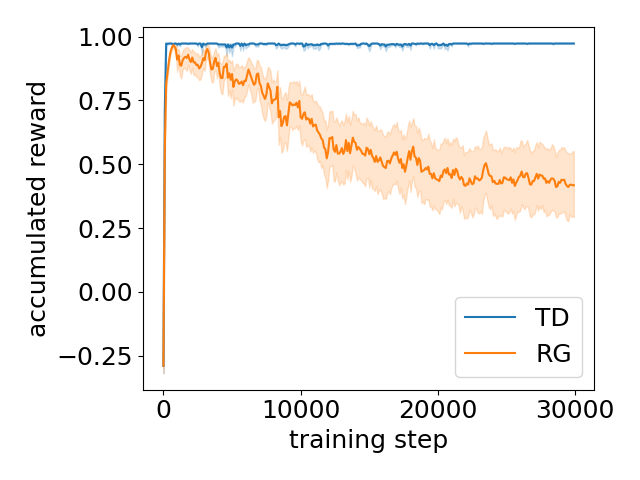}  
  \caption{Grid World}
  \label{fig:sub-first}
\end{subfigure}
\begin{subfigure}{.24\textwidth}
  \centering
  \includegraphics[width=1\linewidth]{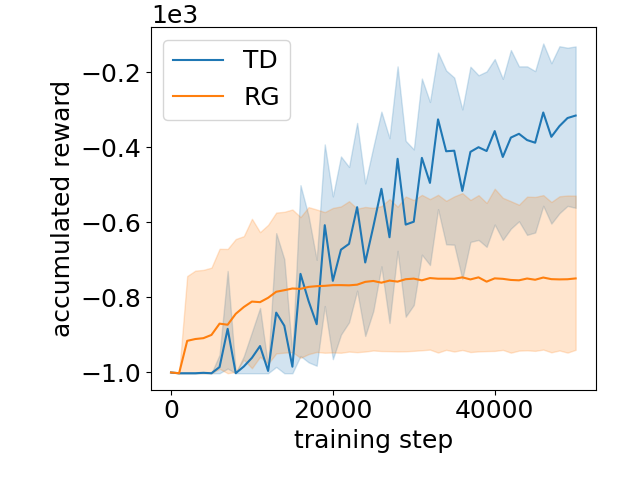}  
  \caption{Acrobot}
  \label{fig:sub-first}
\end{subfigure}
\begin{subfigure}{.24\textwidth}
  \centering
  \includegraphics[width=1\linewidth]{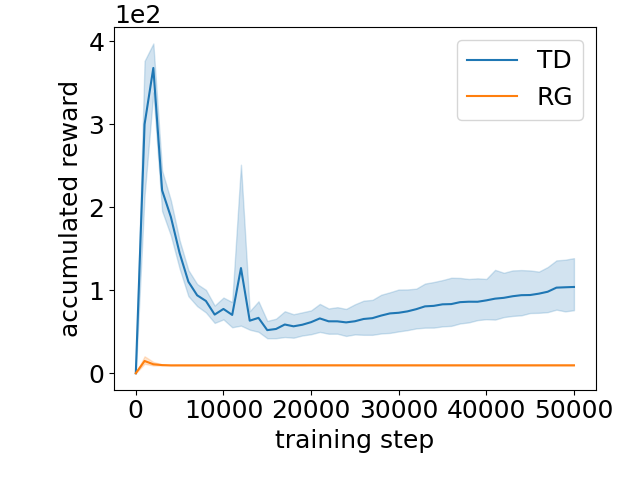}  
  \caption{Cart Pole}
  \label{fig:sub-first}
\end{subfigure}
\begin{subfigure}{.24\textwidth}
  \centering
  \includegraphics[width=1\linewidth]{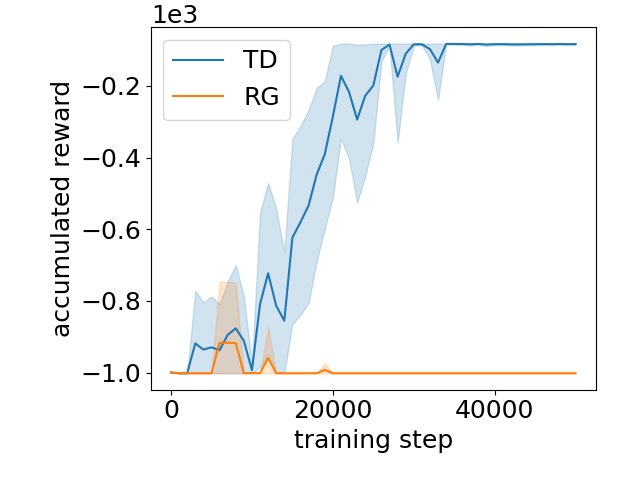}  
  \caption{Mountain Car}
  \label{fig:sub-first}
\end{subfigure}
\caption{Histogram of the accumulated rewards of different environments for TD (blue) and RG (orange). The abscissa represents the training step, the ordinate represents the accumulated reward. Each line represents the mean value of their accumulated rewards and the transparent regions around it represent the $95\%$ confidence interval. (a) contains 80 trials with different initialization, while each of (b, c, d) contains 10 trials. }
\label{fig:discrete state MDP accumulate reward comparsion}
\end{figure}

In this subsection, we aim to compare TD and RG by using the accumulated reward, which is often used as an indicator for performance in practical tasks. We trained two algorithms in the Grid World environment 80 times and in other environments 10 times. In Figure \ref{fig:discrete state MDP accumulate reward comparsion}, the line represents the mean accumulated reward, and the region around it represents the $95\%$ confidence interval. Comparing the result in the figures, the TD method clearly learns a better policy during training in all four environments. Besides, the accumulated reward of RG in Figure \ref{fig:discrete state MDP accumulate reward comparsion} (a) decreases along with training, which indicates policies in some trials learned by RG step into a bad regime, where the loss function contradicts the goodness of policy.

\subsection{Models learned by TD are more robust than RG}

In supervised learning, the learned model is not only concerned with the precision of prediction but also the robustness of the solution. The robustness in the following examples is defined as the decrement of the accumulated reward after parameter perturbation. 

We use the following equation to generate a noisy parameter vector $\vtheta'$ with a noise scale $\alpha$:
\begin{align} \label{eq:noise generation}
    \vtheta' = \vtheta + \alpha \times \mathcal{N}(0, 1),\quad\alpha > 0,
\end{align}

\begin{figure}[htb]
\begin{subfigure}{.24\textwidth}
  \centering
  \includegraphics[width=1\linewidth]{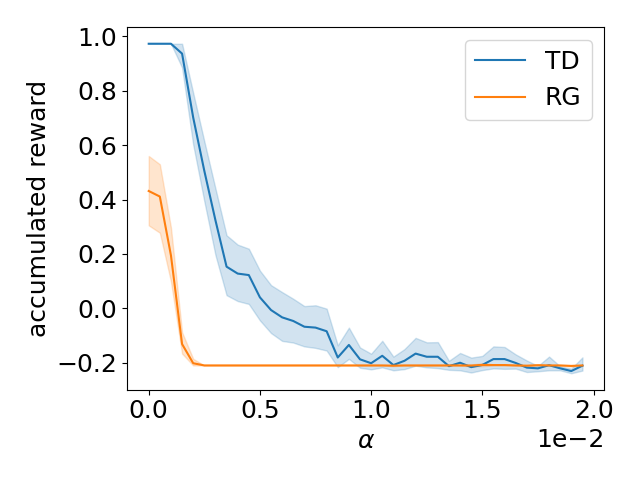}
  \caption{Grid World}
  \label{fig:sub-first}
\end{subfigure}
\begin{subfigure}{.25\textwidth}
  \centering
  \includegraphics[width=1\linewidth]{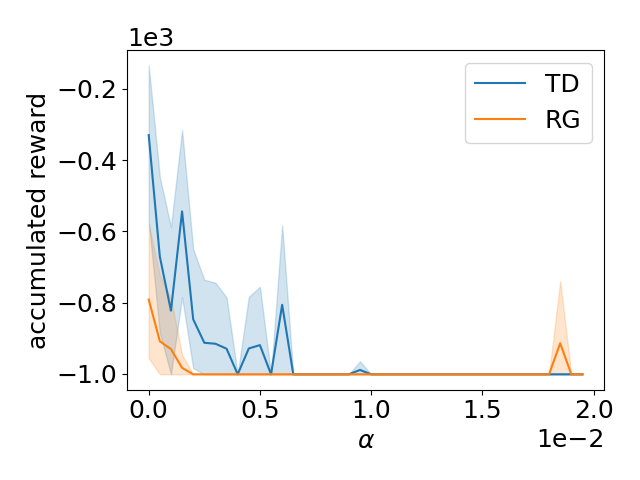}
  \caption{Acrobot}
  \label{fig:sub-first}
\end{subfigure}
\begin{subfigure}{.24\textwidth}
  \centering
  \includegraphics[width=1\linewidth]{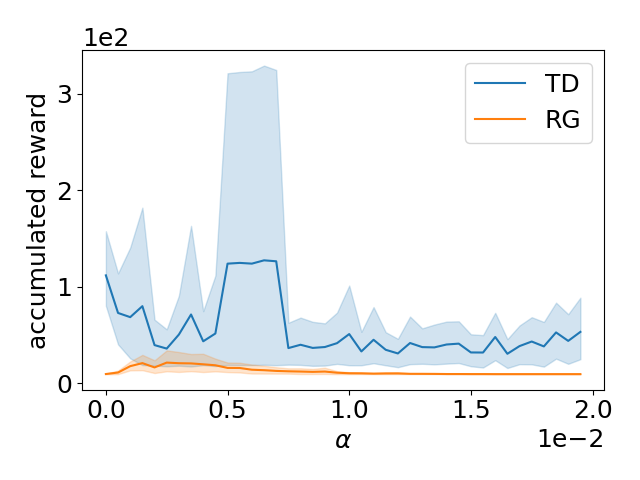}
  \caption{Cart Pole}
  \label{fig:sub-first}
\end{subfigure}
\begin{subfigure}{.25\textwidth}
  \centering
  \includegraphics[width=1\linewidth]{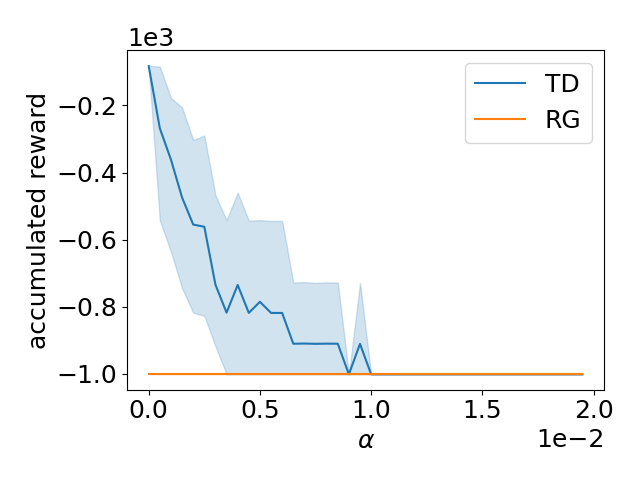}
  \caption{Mountain Car}
  \label{fig:sub-first}
\end{subfigure}
\caption{Performance comparison between TD and RG with perturbation on parameters. In all figures, the abscissa represents the scale $\alpha$ of the perturbation, and the ordinate represents the accumulated reward of the current policy. The noise is obtained by (\ref{eq:noise generation}). At almost all $\alpha$'s, models learned by TD have higher accumulated rewards, that is, TD is more robust.}
\label{fig:Discrete and Continuous MDP robustness}
\end{figure}

In Figure \ref{fig:Discrete and Continuous MDP robustness}, all the policies obtained by TD have higher accumulated rewards than RG at almost all the noise scales. Specifically, in (a), the model learned by TD almost does not change when the perturbation is small, which indicates that the TD method is much more robust in the Grid World environment.

In this section, we conclude that TD can empirically be better and more robust than RG in deep Q-learning. This is the answer to the question of why TD is more popular than RG. Next, we are going to study the special characteristics of deep Q-learning, different from supervised learning.
\section{Analysis of the result with TD and RG}
We assume both the TD method and RG method can achieve a small Bellman residual error. Under this assumption, we have
\begin{align}
    Q(s,a) \approx r(s,a) + \gamma \max_{a' \in A} Q(s',a'),
\end{align}
for any $(s,a,s',r) \in D_N$. If we choose $a = \argmax_{a'' \in A} Q(s,a'')$ and use $V(\cdot)$ to represent $\max_{a \in A} Q(\cdot,a)$, we will have
\begin{align}
    V(s) & \approx r(s,\argmax_{a} Q(s,a)) + \gamma V(s').
\end{align}

\subsection{The RG method can obtain a random policy with a small loss}
In this subsection, we would show that the loss function in deep Q-learning does not indicate the performance of the trained model, which is stark different from supervised learning. 

\begin{figure}[htb]
\begin{subfigure}{.24\textwidth}
  \centering
  \includegraphics[width=1\linewidth]{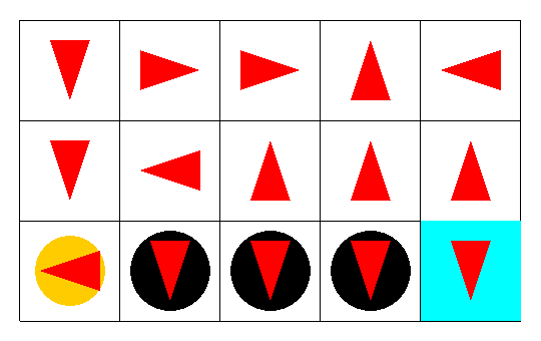}
  \caption{RG's final policy}
  \label{fig:sub-first}
\end{subfigure}
\begin{subfigure}{.24\textwidth}
  \centering
  \includegraphics[width=1\linewidth]{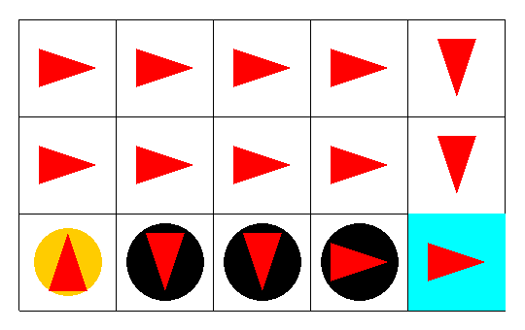}
  \caption{TD's final policy}
  \label{fig:sub-first}
\end{subfigure}
\begin{subfigure}{.24\textwidth}
  \centering
  \includegraphics[width=1\linewidth]{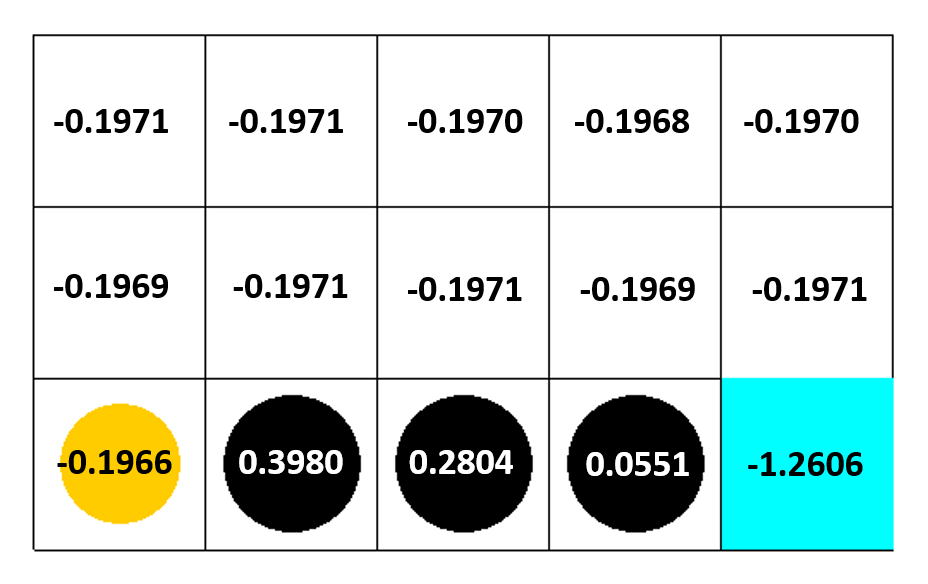}  
  \caption{RG's state value}
  \label{fig:sub-first}
\end{subfigure}
\begin{subfigure}{.24\textwidth}
  \centering
  \includegraphics[width=1\linewidth]{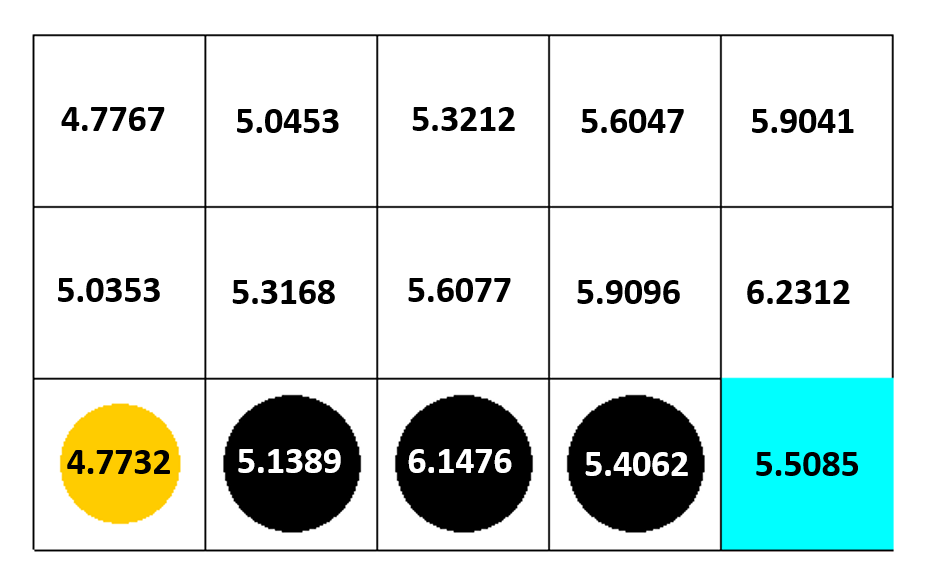}  
  \caption{TD's state value}
  \label{fig:sub-first}
\end{subfigure}
\newline 
\begin{subfigure}{.24\textwidth} \vspace{2mm}
  \centering
  \includegraphics[width=1\linewidth]{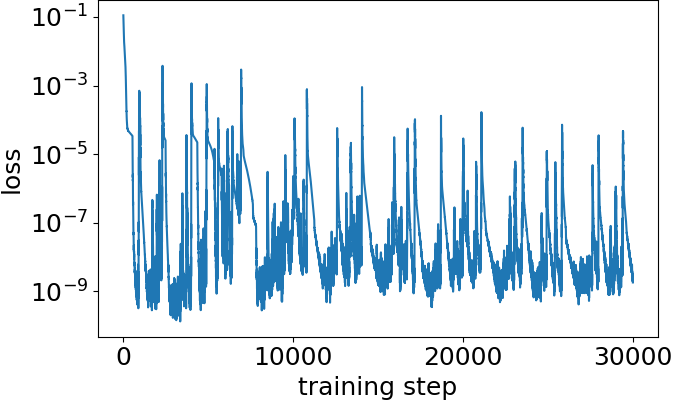}  
  \caption{RG's loss}
  \label{fig:sub-first}
\end{subfigure}
\begin{subfigure}{.24\textwidth} \vspace{2mm}
  \centering
  \includegraphics[width=1\linewidth]{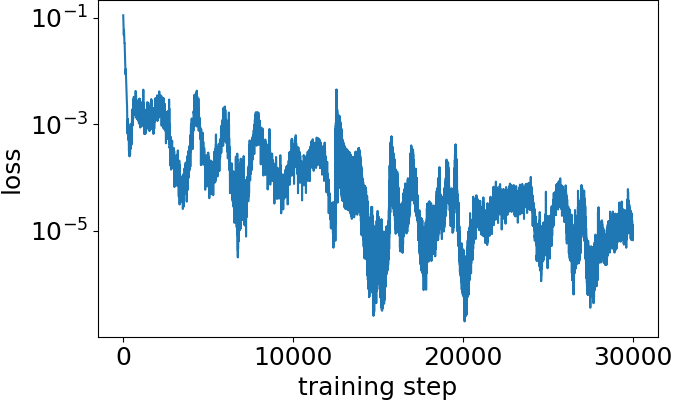}  
  \caption{TD's loss}
  \label{fig:sub-first}
\end{subfigure}
\begin{subfigure}{.24\textwidth} \vspace{2mm}
  \centering
  \includegraphics[width=1\linewidth]{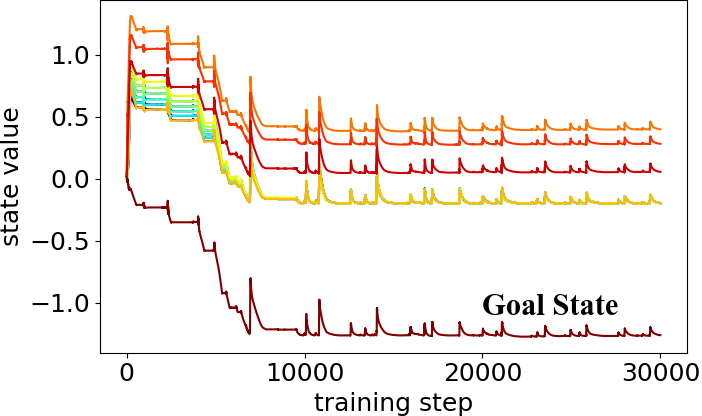}
  \caption{RG's value dynamics}
  \label{fig:sub-first}
\end{subfigure}
\begin{subfigure}{.24\textwidth} \vspace{2mm}
  \centering
  \includegraphics[width=1\linewidth]{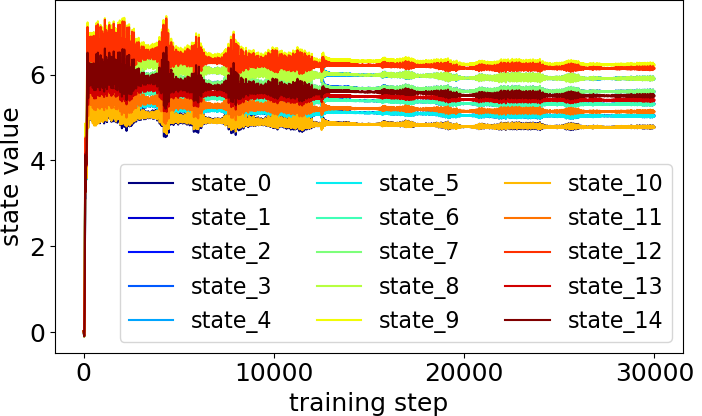}
  \caption{TD's value dynamics}
  \label{fig:sub-first}
\end{subfigure}
\caption{Experiment results comparison between TD and RG in the Grid World problem with the same initial parameter value and training hyper-parameters. Comparing (a, e) with (b, f), RG's loss is around $10^{-9}$ and RG learns a bad policy, TD's loss is around $10^{-5}$ and TD learns a good policy. This means lower loss does not indicate a better policy. In (c, g), state value is in the bad regime. However, in (d, h), state value is in the good regime.  Besides, in (d), because the goal state has a reward $+1$, it can have a lower value than the previous state.}
\label{fig:grid world learned policy}
\end{figure}

Figure \ref{fig:grid world learned policy} (a) shows the policy learned by RG, the red triangle represents the action with maximum state-action value. Clearly, (a) shows a bad policy. However, the loss in Figure \ref{fig:grid world learned policy} (e) is around $10^{-9}$, which is a very small number. By comparing the good policy learned by TD in Figure \ref{fig:grid world learned policy} (b) and the loss around $10^{-5}$ in (f), it is easy to notice that lower Bellman residual error does not indicate a better policy. 
The following analysis will show that there are distinct regimes of the state value where the Bellman residual can all be small but the performance of policy can be very different.   

Intuitively speaking, for each non-terminal state of the Grid World problem, we want there exists a path from the state to the pre-terminal state of the goal, where the state value monotonically increases and the pre-terminal state is the state before the terminal state. This means if a state is closer to the goal, it should have a higher state value.  For a pre-terminal state $s$ and corresponding terminal state $s'$, we have
\begin{align}
    Q(s,a) = r_T(s,a) + \gamma \max_{a' \in A} Q(s',a').
\end{align}
Since $r_T(s,a)$ is positive, it does not have to require the highest state value for the terminal state. Similarly, the policy in the Mountain Car and Acrobot problem follows the same analysis because the goal of these two problems is to go to the terminal states similar to the Grid World. Mathematically, this means given a data tuple $z = (s,a,s',r)$ , where $a = \argmax_{a' \in A} Q(s.a)$ and $s'$ is closer to the goal but not a terminal state, we want the following inequality holds
\begin{align}
    V(s) & \approx r_s(s,\argmax_{a} Q(s,a)) + \gamma V(s') < V(s') \\
    & \Rightarrow V(s') > \frac{r_s(s,\argmax_{a} Q(s,a))}{1-\gamma}.
\end{align}
For example, we consider the Grid World problem, we have $\gamma = 0.95$, $r_s=-0.01$, then, we obtain
\begin{equation}
    V(s') > \frac{-0.01}{1-0.95} = -0.2.
\end{equation} 
To ensure the monotonicity, it requires all the state values except for the terminal state and the initial state greatly larger than $-0.2$, that is, the regime for a good policy. Otherwise, it is a bad regime that Bellman residual is small but the policy is bad. Figure \ref{fig:grid world learned policy} (c) shows an RG example in the bad regime. For all  states except terminal states, the value is around $-0.2$ in the bad regime with a small Bellman residual error shown in (e) and a bad policy shown in (a). However, for the TD case in (d), all the non-terminal state values are greatly larger than $-0.2$, which indicates this value function is in a good policy regime. Consistent with theoretical analysis, we obtain a good policy in (b). 

We then look into how these state values evolve during the training process for the two training methods. For the RG method, as shown in Figure \ref{fig:grid world learned policy} (g), the goal state almost monotonically decreases with the training step and achieves -1.26, the lowest value among all state values. When the Bellman equation holds, all the other non-terminal state values can be computed as -0.2.  On the contrary, for the TD method, as shown in Figure \ref{fig:grid world learned policy} (h), all state value functions increase with the training step initially. Taken together, studying the dynamic behavior may be a key to understanding why these two algorithms take the same initial values to different regimes. 
Statistically, we conduct 80 trials in the Grid World environment with different initialization, and we find that 38 out of 80 times the RG method learns a bad policy and TD learns all good policies. 

\begin{figure}[htb]
\begin{subfigure}{.24\textwidth}
  \centering
  \includegraphics[width=1\linewidth]{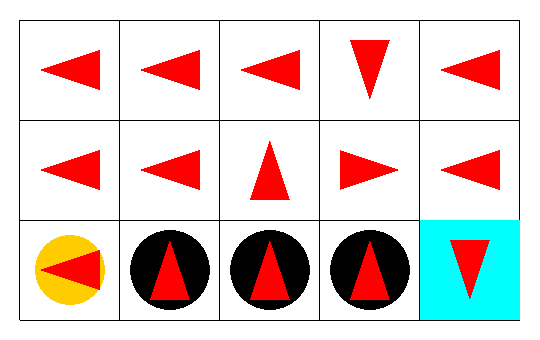}
  \caption{RG's final policy}
  \label{fig:sub-first}
\end{subfigure}
\begin{subfigure}{.24\textwidth}
  \centering
  \includegraphics[width=1\linewidth]{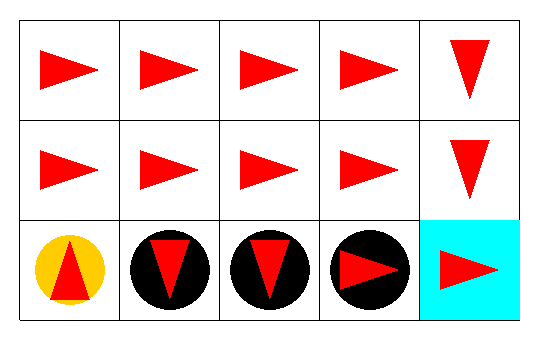}
  \caption{TD's final policy}
  \label{fig:sub-first}
\end{subfigure}
\begin{subfigure}{.24\textwidth}
  \centering
  \includegraphics[width=1\linewidth]{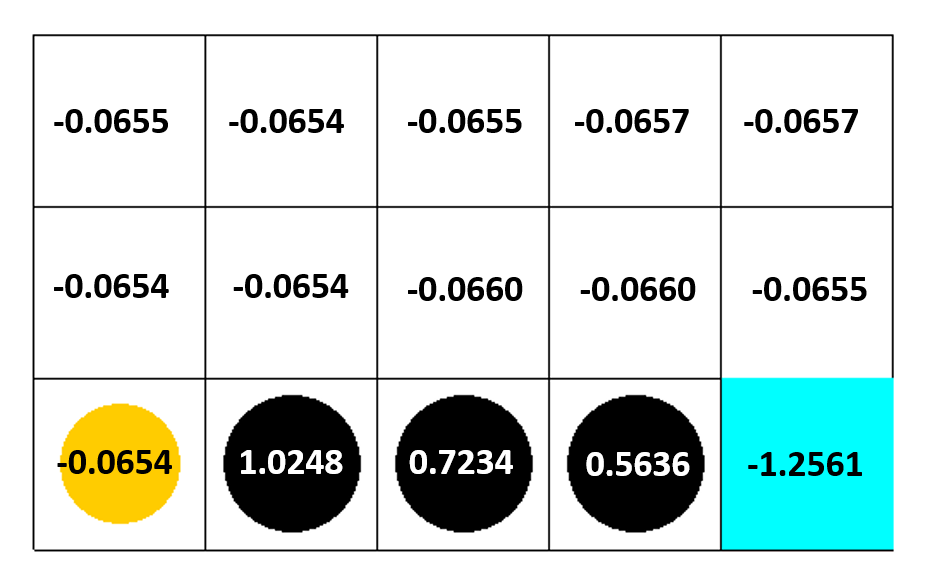}  
  \caption{RG's state value}
  \label{fig:sub-first}
\end{subfigure}
\begin{subfigure}{.24\textwidth}
  \centering
  \includegraphics[width=1\linewidth]{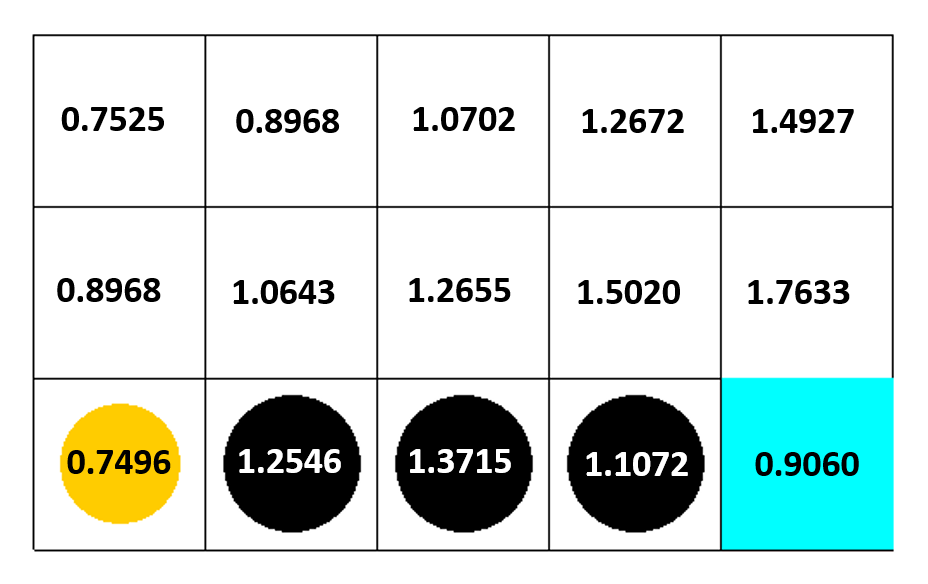}  
  \caption{TD's state value}
  \label{fig:sub-first}
\end{subfigure}
\newline 
\begin{subfigure}{.24\textwidth} \vspace{2mm}
  \centering
  \includegraphics[width=1\linewidth]{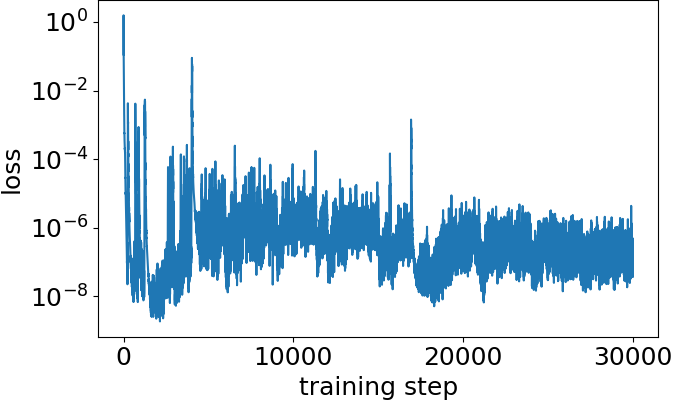}  
  \caption{RG's loss}
  \label{fig:sub-first}
\end{subfigure}
\begin{subfigure}{.24\textwidth} \vspace{2mm}
  \centering
  \includegraphics[width=1\linewidth]{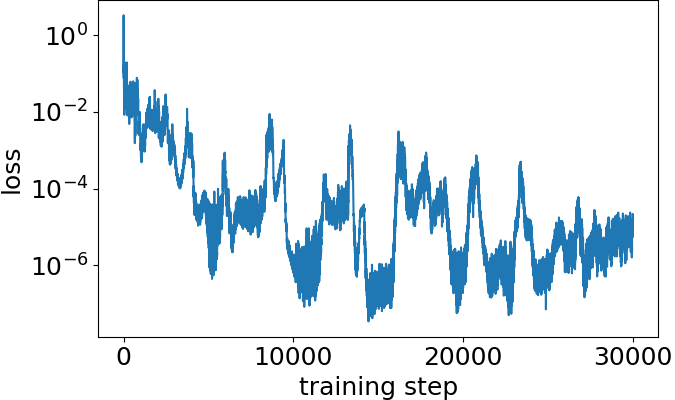}  
  \caption{TD's loss}
  \label{fig:sub-first}
\end{subfigure}
\begin{subfigure}{.24\textwidth} \vspace{2mm}
  \centering
  \includegraphics[width=1\linewidth]{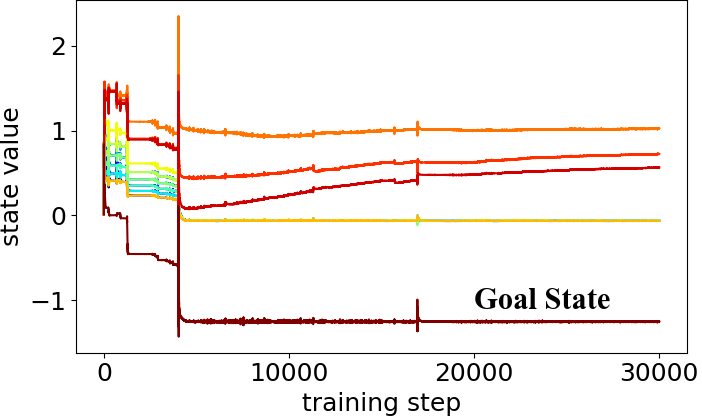}
  \caption{RG's value dynamics}
  \label{fig:sub-first}
\end{subfigure}
\begin{subfigure}{.24\textwidth} \vspace{2mm}
  \centering
  \includegraphics[width=1\linewidth]{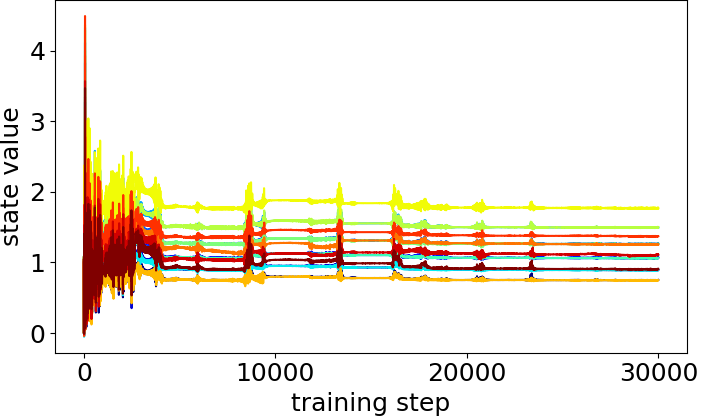}
  \caption{TD's value dynamics}
  \label{fig:sub-first}
\end{subfigure}
\caption{Experiment results comparison between TD and RG in the Grid World problem with $\gamma=0.85$. To increase the learning speed, we increase the initial learning rate to 0.001. In (c), all the states other than the terminal state reach a value around $r_t / (1-\gamma)=-0.066$, which is a bad policy regime. In (h), all the states are larger than 0.066, which is a good policy regime. The argument of policy regime holds for different $\gamma$.}
\label{fig:grid world learned policy gamma=0.85}
\end{figure}

\begin{figure}[htb]
\begin{subfigure}{.24\textwidth}
  \centering
  \includegraphics[width=1\linewidth]{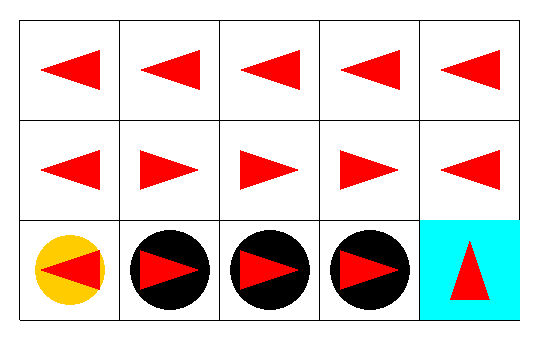}
  \caption{RG's final policy}
  \label{fig:sub-first}
\end{subfigure}
\begin{subfigure}{.24\textwidth}
  \centering
  \includegraphics[width=1\linewidth]{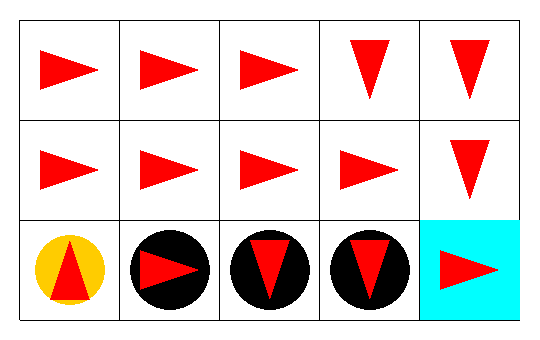}
  \caption{TD's final policy}
  \label{fig:sub-first}
\end{subfigure}
\begin{subfigure}{.24\textwidth}
  \centering
  \includegraphics[width=1\linewidth]{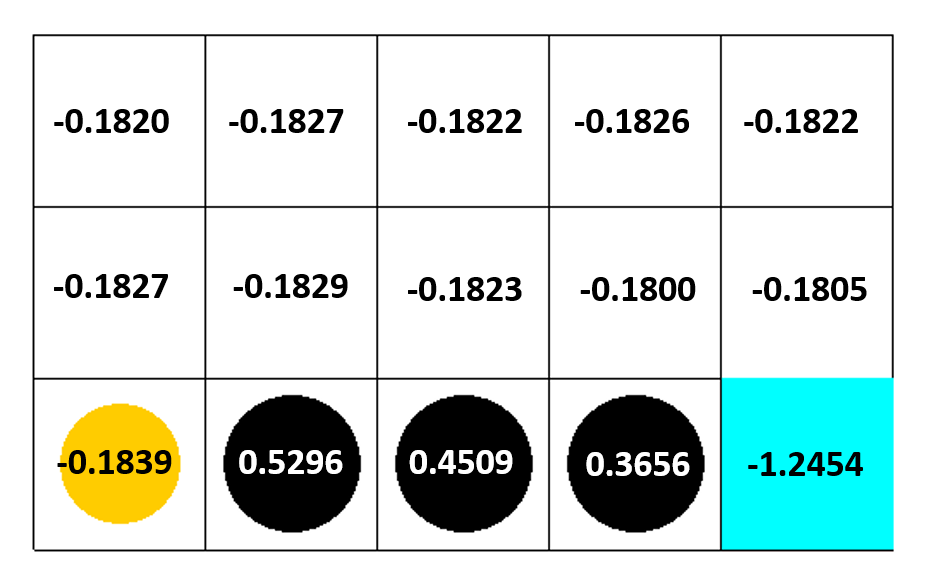}  
  \caption{RG's state value}
  \label{fig:sub-first}
\end{subfigure}
\begin{subfigure}{.24\textwidth}
  \centering
  \includegraphics[width=1\linewidth]{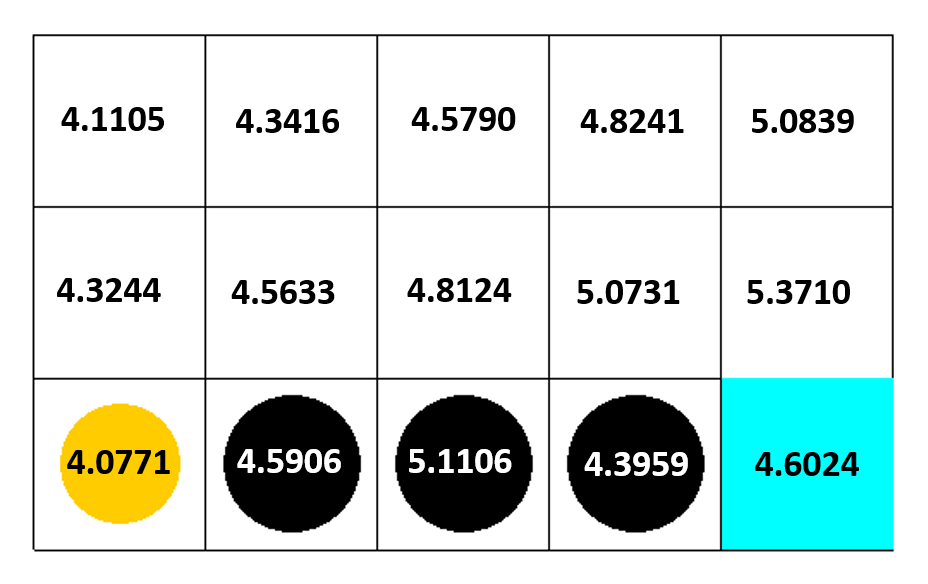}  
  \caption{TD's state value}
  \label{fig:sub-first}
\end{subfigure}
\newline 
\begin{subfigure}{.24\textwidth} \vspace{2mm}
  \centering
  \includegraphics[width=1\linewidth]{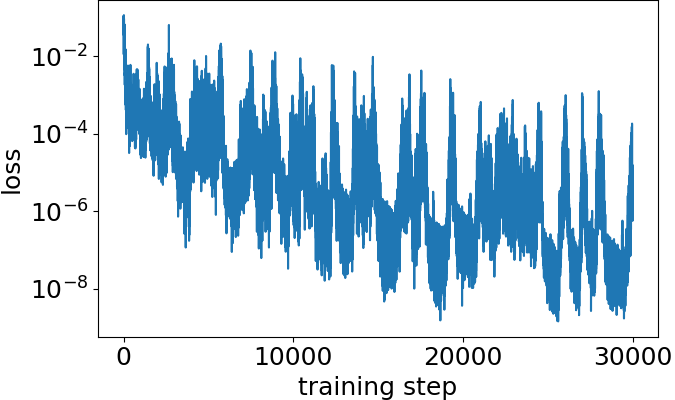}  
  \caption{RG's loss}
  \label{fig:sub-first}
\end{subfigure}
\begin{subfigure}{.24\textwidth} \vspace{2mm}
  \centering
  \includegraphics[width=1\linewidth]{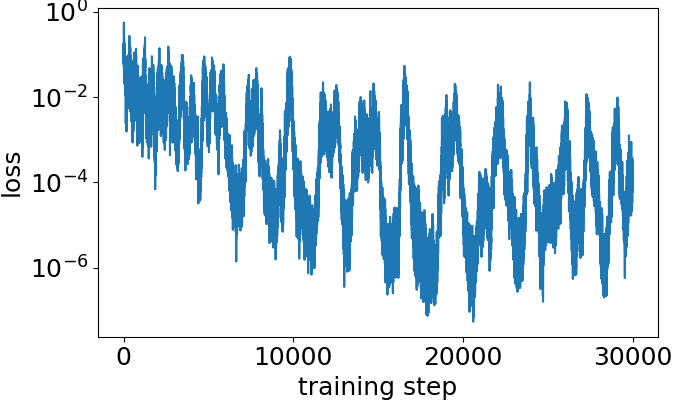}  
  \caption{TD's loss}
  \label{fig:sub-first}
\end{subfigure}
\begin{subfigure}{.24\textwidth} \vspace{2mm}
  \centering
  \includegraphics[width=1\linewidth]{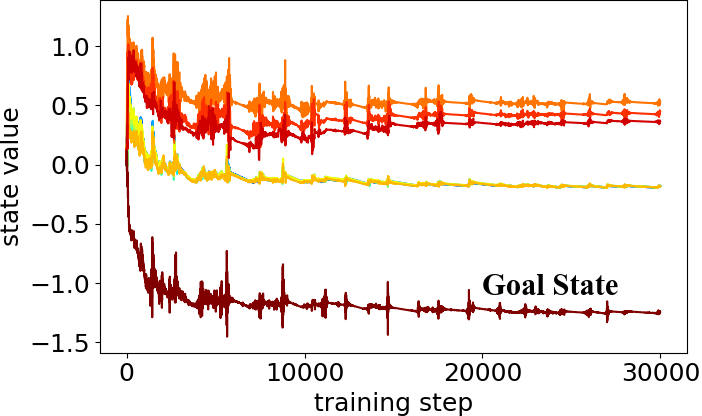}
  \caption{RG's value dynamics}
  \label{fig:sub-first}
\end{subfigure}
\begin{subfigure}{.24\textwidth} \vspace{2mm}
  \centering
  \includegraphics[width=1\linewidth]{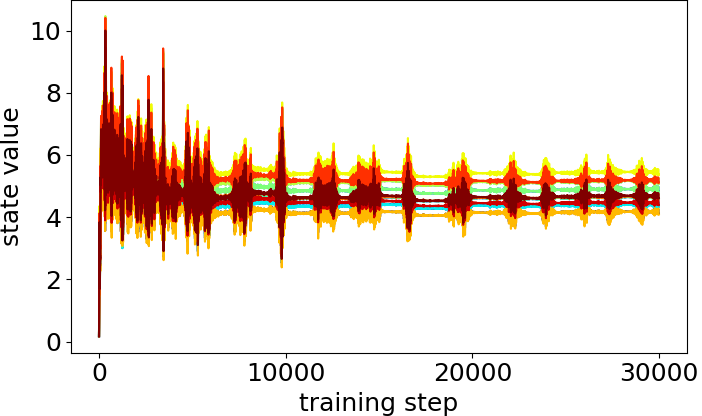}
  \caption{TD's value dynamics}
  \label{fig:sub-first}
\end{subfigure}
\caption{Experiment results comparison between TD and RG in the Grid World problem with on-policy, mini-batch gradient descent, and experience replay. We set the batch size to 16, buffer size to 30000, and random exploration rate to 0.2. To speed up training, we set the initial learning rate to 0.0005. The result is similar to Figure \ref{fig:grid world learned policy}. The argument for a policy regime still holds.}
\label{fig:grid world learned policy online}
\end{figure}

To extend our result to more general settings, we conduct two more experiments. In figure \ref{fig:grid world learned policy gamma=0.85}, we demonstrated that even if the $\gamma$ changes, the analysis of the policy regime still holds. In figure \ref{fig:grid world learned policy online}, we demonstrated that the understanding of the policy regime can also apply to the general DQN setting, which is on-policy, mini-batch gradient descent, and experience replay. 

\begin{figure}[htb]
\centering
\begin{subfigure}[t]{.24\textwidth}
  \centering
  \includegraphics[width=1\linewidth]{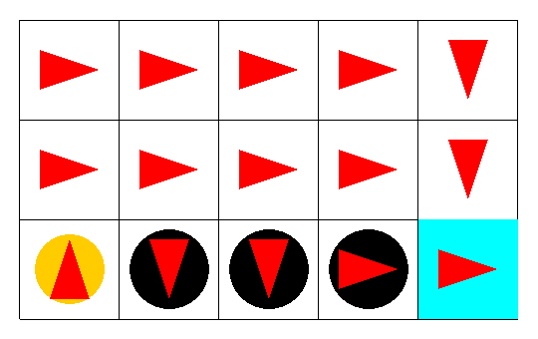}
  \caption{final policy}
  \label{fig:sub-first}
\end{subfigure}
\begin{subfigure}[t]{.24\textwidth}
  \centering
  \includegraphics[width=1\linewidth]{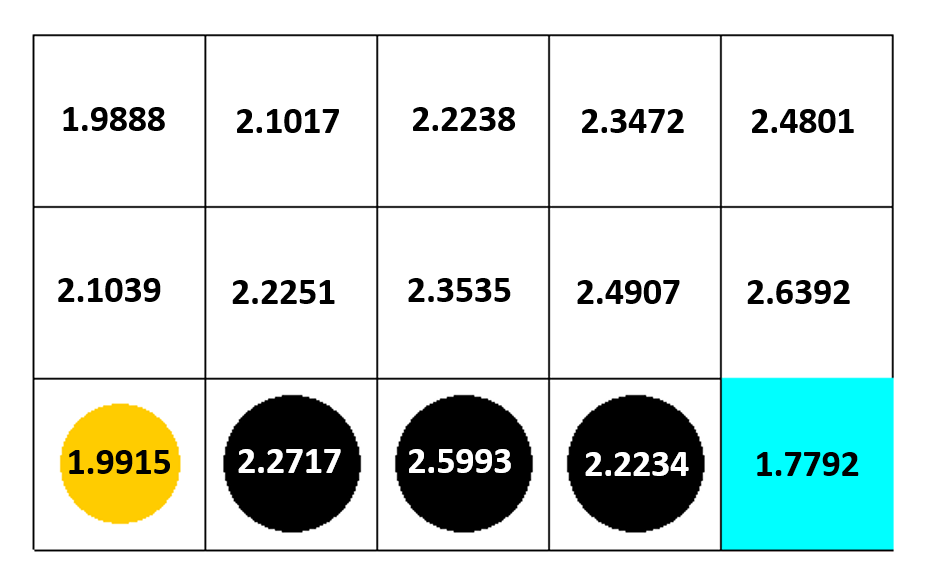}
  \caption{final state value}
  \label{fig:sub-first}
\end{subfigure}
\begin{subfigure}[t]{.24\textwidth}
  \centering
  \includegraphics[width=1\linewidth]{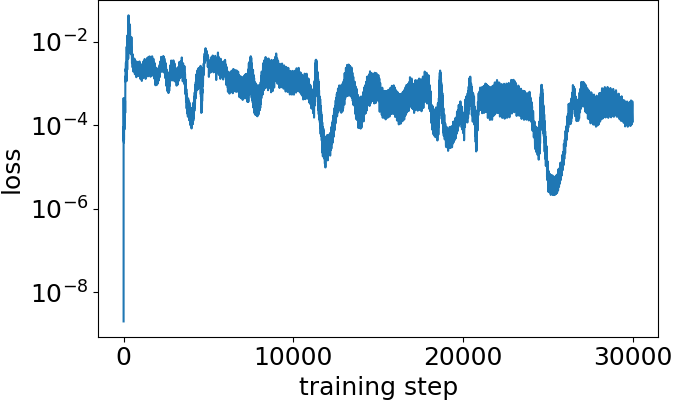}
  \caption{loss}
  \label{fig:sub-first}
\end{subfigure}
\begin{subfigure}[t]{.24\textwidth}
  \centering
  \includegraphics[width=1\linewidth]{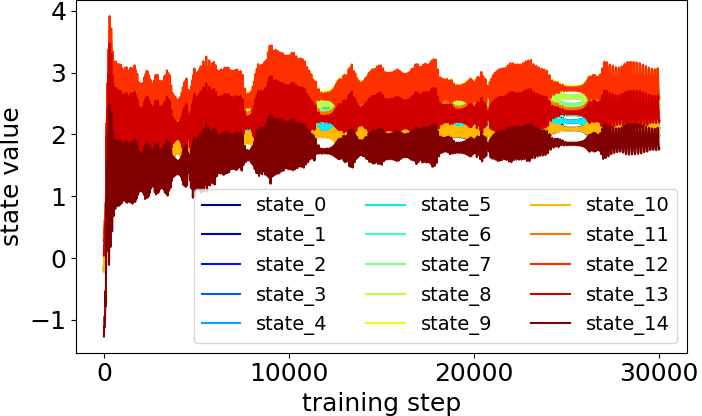}
  \caption{state value dynamics}
  \label{fig:sub-first}
\end{subfigure}
\caption{The result of the model that is continuously trained by TD from the model trained by RG in Figure \ref{fig:grid world learned policy} (a). (a, b) show the policy and state value after using TD  to train for 30000 steps. As we can see, it jumps out of the bad policy and learns a good policy. (c, d) show the loss and state value dynamics during training. There is an impulse at the beginning of these two figures, which could represents jumping out of old solution.}
\label{fig:training switch from RG to TD}
\end{figure}

In addition, to show the superiority of TD compared to RG, we use the model learned by RG as initialization and train this model with the TD method for 30000 steps. As an example shown in Figure \ref{fig:training switch from RG to TD} (a, b), TD learns a good policy and state values satisfy the monotonicity in general. The TD method jumps out of the bad policy regime learned by the RG method and enters the good policy regime where all the state values are greatly larger than -0.2. Figure \ref{fig:training switch from RG to TD} (c, d) represents the loss and state value dynamics during training. At the beginning of these two figures, the loss and state value increase dramatically, which may represent the neural network model jumping out of the bad policy regime.

\subsection{In Cart Pole, the RG method learns a bad policy}

The Cart Pole problem is different from the Grid World problem, which is trying to prevent the agent from going to terminal states. Therefore, the regime of good policies is also different. In order to simplify the following discussion, we define the terminal state set as 
$S_T = \{s| \text{ for all } s \in S \text{ and } s \text{ is terminal state. } \},  $
the pre-terminal state set as
$
  S_{pT} = \{s| s \in (s,\argmax_{a' \in A} Q(s.a),s',r) \text{ where } s' \in S_T \}, $
and the other state set as
$
    S_{O} = S \backslash (S_T \cup S_{pT}).
$

We also define the mean state value of the terminal state as
$
    V_T = \frac{1}{|S_T|}\sum_{ s\in S_T} \max_{a \in A} Q(s,a),   
$
mean state value of pre-terminal state value as
$
    V_{pT} = \frac{1}{|S_{pT}|}\sum_{ s\in S_{pT}} \max_{a \in A} Q(s,a),
$
and mean state value of the other state set as
$
    V_O = \frac{1}{|S_O|}\sum_{ s\in S_O} \max_{a \in A} Q(s,a),    
$
where $|S|$ represents the element number of set $S$.

One way to keep the agent alive is to let the agent move in a loop. To construct a loop, the state value in the loop should not increase or decrease monotonically, all the state values should be equal. This means for a given data tuple $(s, \argmax_{a} Q(s,a), s', r)$ in the loop, we should have
\begin{align}
    V(s) & \approx r_s(s,\argmax_{a} Q(s,a)) + \gamma V(s') = V(s') \\
    & \Rightarrow V(s') = \frac{r_s(s,\argmax_{a} Q(s,a))}{1-\gamma} = \frac{1}{1-0.87} \approx 7.6923.
\end{align}
Also, we want for all $s \in S_{pT}$  to have $V(s) \ll 7.6923$, so the agent does not easily go to these states. Therefore, a small terminal state value is preferred according to Bellman equation.
However, the Cart Pole problem $S_O$ may contain states that are not in the loop. To ensure the agent does not leave the loop, those states in $S_O$ and not in the loop should have lower state values, smaller than $7.6923$, leading to $V_O<7.6923$. Also, we want $V_{pT} \ll 7.6923$ and a small $V_T$. This is a good regime of policies, otherwise, it is highly likely a bad regime.

\begin{figure}[htb]
\centering
\begin{subfigure}{.31\textwidth}
  \centering
  \includegraphics[width=1\linewidth]{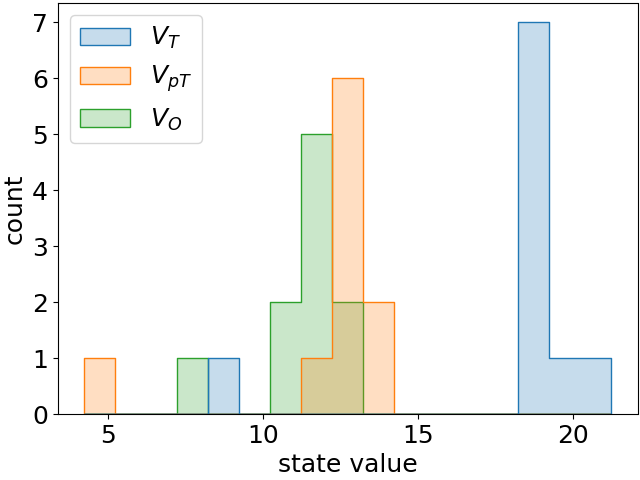}  
  \caption{RG} 
  \label{fig:sub-first}
\end{subfigure} \hspace{3mm}
\begin{subfigure}{.31\textwidth}
  \centering
  \includegraphics[width=1\linewidth]{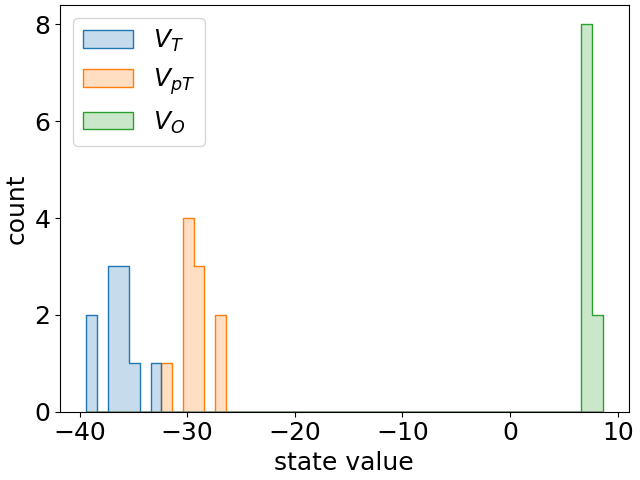}  
  \caption{TD}
  \label{fig:sub-first}
\end{subfigure} \hspace{1mm}
\begin{subfigure}{.31\textwidth}
  \centering
  \includegraphics[width=1\linewidth]{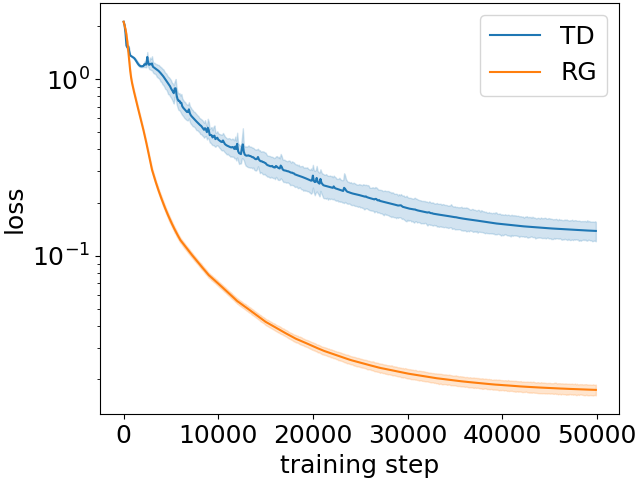}  
  \caption{loss}
  \label{fig:sub-first}
\end{subfigure}
\caption{Histogram of $V_O$, $V_{pT}$ and $V_T$ for ten experiments in the Cart Pole problem of TD and RD training and the training loss for ten experiments. In (a), $V_{pT}$ and $V_T$ obtained by RG are larger than 7.6923 for nine times, so the agent will go to the terminal state following the increasing state values.  On the contrary in (b), $V_{pT}$ and $V_T$ obtained by TD are pretty small, all around -30, so the agent will not go to the terminal state easily. In (c), the line represents the mean loss, and the region around it is the $95\%$ confidence interval. The mean loss for TD is $0.138$ and $0.017$ for RG. Same as the Grid World problem, RG learns a bad policy with a lower loss. Too large terminal state value in the Cart Pole problem indicates the model learned by RG achieves a bad policy.}
\label{fig:Cart Pole Learning Dynamics}
\end{figure}

In Section \ref{sec:TDgoodexp}, we have shown that RG always learns a bad policy in the Cart Pole problem in Figure \ref{fig:discrete state MDP accumulate reward comparsion} (c). Besides, RG also has a lower loss shown in Figure \ref{fig:Cart Pole Learning Dynamics} (c). This is because the model learned by RG falls into a bad regime of policies. Figure \ref{fig:Cart Pole Learning Dynamics} shows $V_O$, $V_{pT}$ and $V_{T}$ of RG and TD for all ten experiments in the Cart Pole problem. In Figure  \ref{fig:Cart Pole Learning Dynamics} (a), nine of ten times $V_{pT}$ and $V_T$ obtained by RG is larger than 7.6923. Although the remaining one has $V_{pT} < 7.6923$, we find that most state values are in a bad regime. For example, in 100 random play sequence data, we count the pre-terminal state value and terminal state value of each play predicted by the trained model. We find that 24 and 66 times the pre-terminal state values and the terminal state values are larger than $7.6923$ in 100 random play sequence data, respectively. Too large terminal state value in the Cart Pole Problem indicates RG learns a bad policy. On the contrary, for all the models learned by TD shown in \ref{fig:discrete state MDP accumulate reward comparsion} (b), $V_{pT}$ are around -30, $V_T$ are less than -30, which is a pretty small number, and $V_{O}$ are around 7. Models learned by TD are in the good policy regime.

\begin{figure}[htb]
\centering
\begin{subfigure}{.24\textwidth}
  \centering
  \includegraphics[width=1\linewidth]{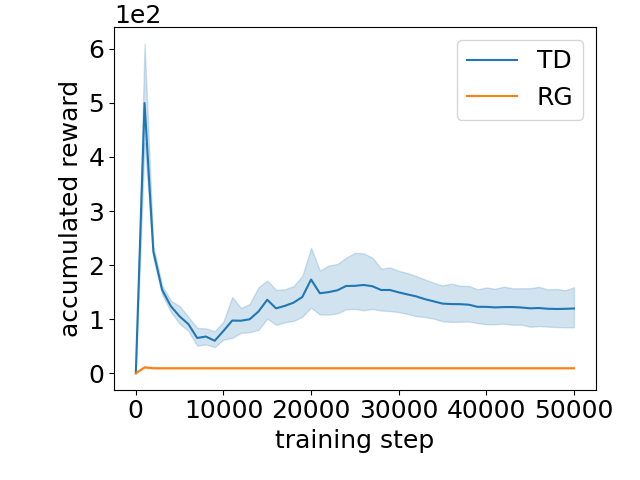}  
  \caption{Accumulate Reward} 
  \label{fig:sub-first}
\end{subfigure} 
\begin{subfigure}{.24\textwidth}
  \centering
  \includegraphics[width=1\linewidth]{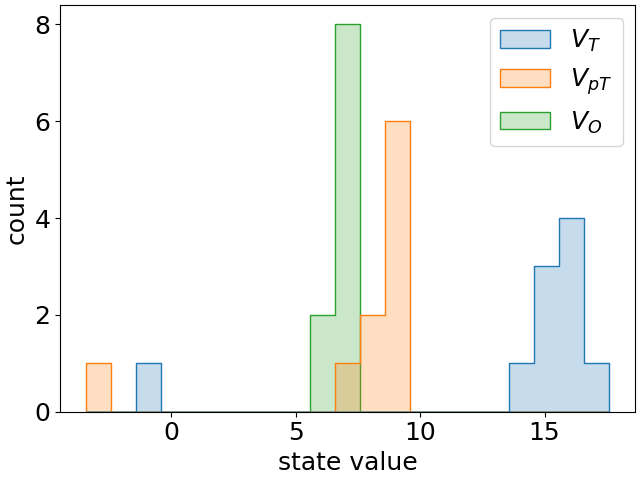}  
  \caption{RG}
  \label{fig:sub-first}
\end{subfigure} 
\begin{subfigure}{.24\textwidth}
  \centering
  \includegraphics[width=1\linewidth]{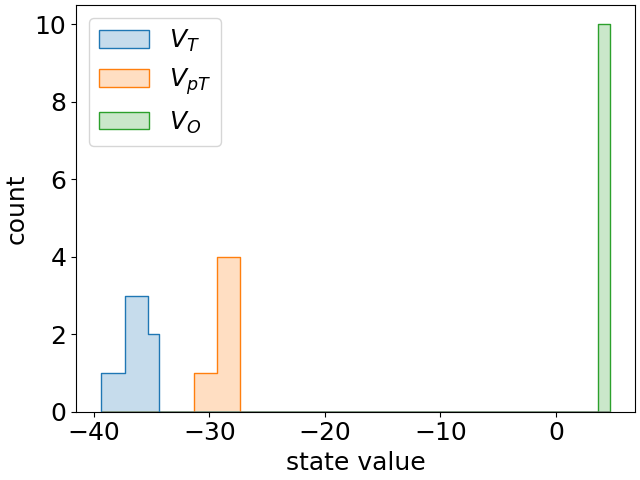}
  \caption{TD}
  \label{fig:sub-first}
\end{subfigure}
\begin{subfigure}{.24\textwidth}
  \centering
  \includegraphics[width=1\linewidth]{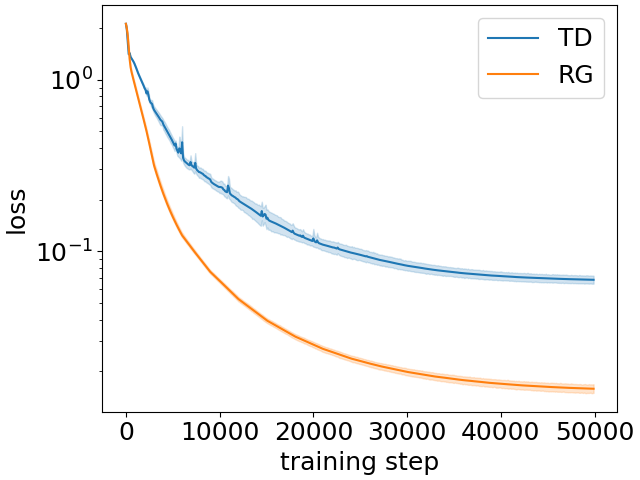}  
  \caption{loss}
  \label{fig:sub-first}
\end{subfigure}
\caption{Histogram of $V_O$, $V_{pT}$ and $V_T$ for ten experiments in the Cart Pole problem of TD and RD with $\gamma=0.8$. When $\gamma=0.8$, the good policy regime is define as: $V_O$ around $r_t/(1-\gamma)=5$ and $V_{pT}$ far smaller than 5. In (c), $V_O$ is actually around 4, this happens because the loss is not small enough, but $V_{pT}$ in (c) is small enough. In (b), however, nine of ten $V_{pT}$ larger than 5, the last one smaller than 5 because some of the pre-terminal state value is pretty small but most of the value is larger than 5. So the argument of policy regime holds for different $\gamma$.}
\label{fig:Cart Pole Learning Dynamics gamma=0.8}
\end{figure}

\begin{figure}[htb]
\centering
\begin{subfigure}{.24\textwidth}
  \centering
  \includegraphics[width=1\linewidth]{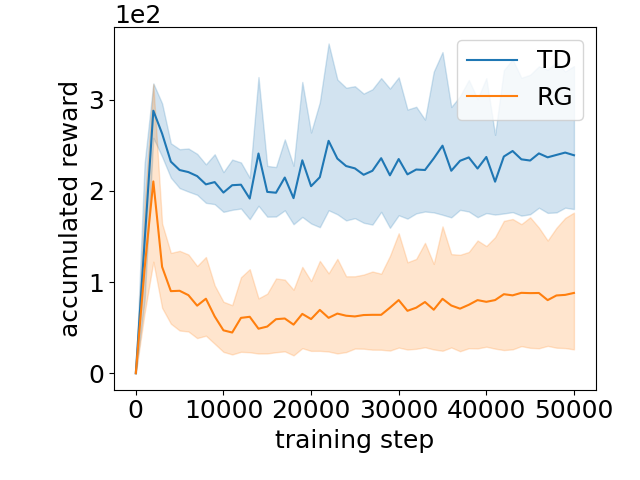}  
  \caption{Accumulate Reward} 
  \label{fig:sub-first}
\end{subfigure} 
\begin{subfigure}{.24\textwidth}
  \centering
  \includegraphics[width=1\linewidth]{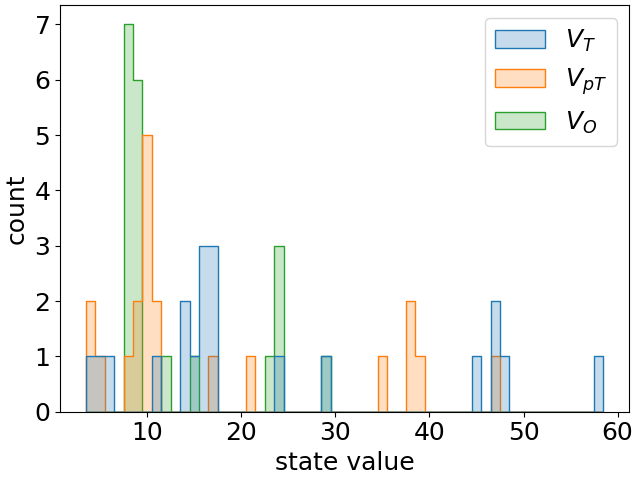}  
  \caption{RG}
  \label{fig:sub-first}
\end{subfigure} 
\begin{subfigure}{.24\textwidth}
  \centering
  \includegraphics[width=1\linewidth]{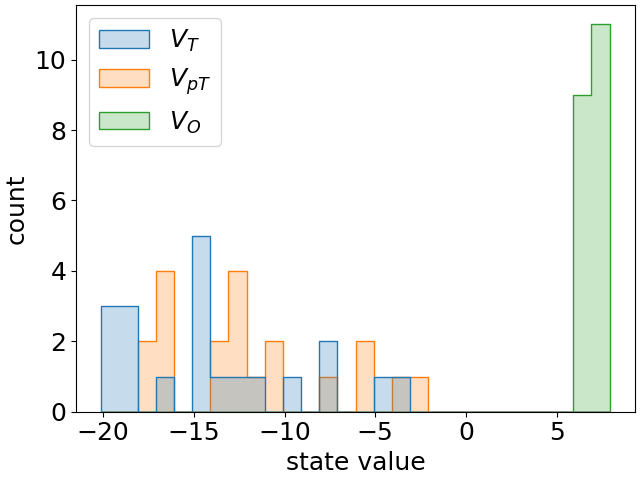}
  \caption{TD}
  \label{fig:sub-first}
\end{subfigure}
\begin{subfigure}{.24\textwidth}
  \centering
  \includegraphics[width=1\linewidth]{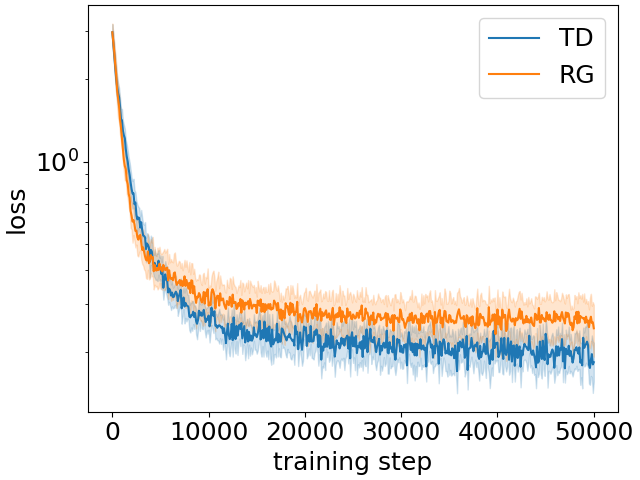}  
  \caption{loss}
  \label{fig:sub-first}
\end{subfigure}
\caption{Histogram of $V_O$, $V_{pT}$ and $V_T$ for twenty experiments in the Cart Pole problem of TD and RD training with on-policy, mini-batch gradient descent, and experience replay. We set the batch size to 1024, buffer size to 50000, and random exploration rate to 0.2. In (c), all the $V_O$ is around 6 and 7, $V_{pT}$ all smaller than $r_t/(1-\gamma) \approx 7.6923$, which is in good policy regime. In (b), there are three times $V_O$ around 7 and $V_{pT} < 7.6923$, so the average performance of RG is better in  an offline setting as shown in (a). The argument of policy regime still holds for an online settings.}
\label{fig:Cart Pole Learning Dynamics online}
\end{figure}

To extend our result to more general settings, we conduct two more experiments. In figure \ref{fig:Cart Pole Learning Dynamics gamma=0.8}, we demonstrated that even if the $\gamma$ changes, the analysis of the policy regime still holds. In figure \ref{fig:Cart Pole Learning Dynamics online}, we demonstrated that the understanding of the policy regime can also apply to the general DQN setting, which is on-policy, mini-batch gradient descent, and experience replay. 

\section{Learning dynamics of backward-semi gradient}

\begin{figure}[htb]
\centering
\begin{subfigure}[t]{.32\textwidth}
  \centering
  \includegraphics[width=1\linewidth]{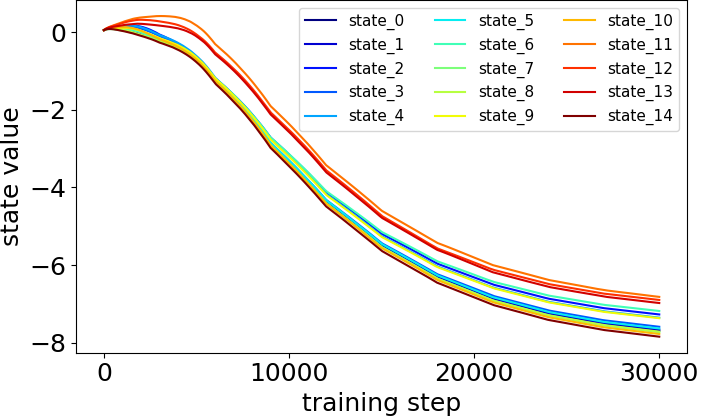}  
  \caption{state value dynamics}
  \label{fig:sub-first}
\end{subfigure}
\begin{subfigure}[t]{.32\textwidth}
  \centering
  \includegraphics[width=1\linewidth]{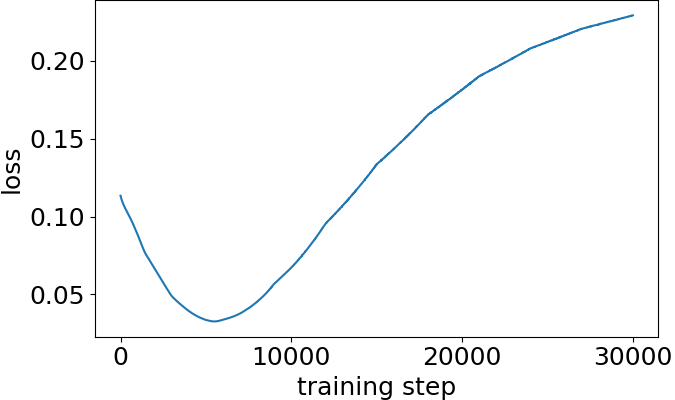} 
  \caption{loss}
  \label{fig:sub-first}
\end{subfigure} 
\begin{subfigure}[t]{.32\textwidth}
  \centering
  \includegraphics[width=1\linewidth]{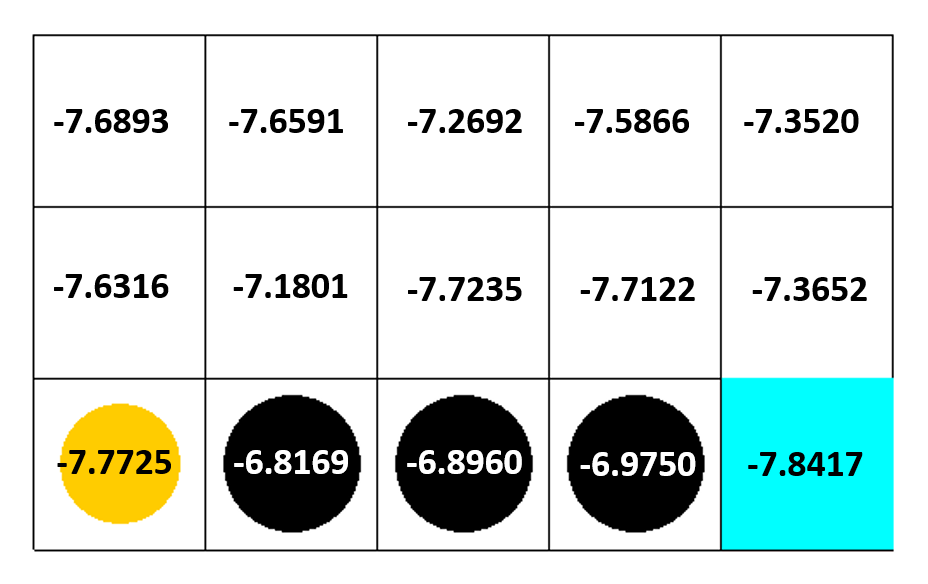} 
  \caption{final state value}
  \label{fig:sub-first}
\end{subfigure} 
\caption{State values are obtained by using $\nabla \mathcal{L}_{\text{backward-semi}}$ as a gradient to train the neural network model in the Grid World environment. (a) shows all the state values decreasing along with training. (b) shows the training loss and it is increasing in the latter stage, which means $\nabla \mathcal{L}_{\text{backward-semi}}$ does not minimize Bellman residual error. (c) shows that the three penalty states, which have a negative reward, have the top three values among all the states, and the goal state, which has a positive reward, has the lowest value. }
\label{fig:backward-semi learning dynamics}
\end{figure}

The mathematical difference between TD and RG is their gradient form. We define the difference between $\nabla \mathcal{L}_{\text{true}}$ and $\nabla \mathcal{L}_{\text{semi}}$ as backward-semi gradient
\begin{align}
    \nabla \mathcal{L}_{\text{backward-semi}} 
    & = \nabla \mathcal{L}_{\text{true}} - \nabla \mathcal{L}_{\text{semi}} \\
     & = \frac{2}{N} \sum_{i=1}^N (Q_{\vtheta}(s_i,a_i) - r(s_i, a_i) -\gamma \max_{a' \in A} Q_{\vtheta}(s'_i, a'))(- \gamma \nabla \max_{a' \in A} Q_{\vtheta}(s'_i, a')).
\end{align}
We used the backward-semi gradient to update the model parameters in the Grid World environment. To ensure the convergence, we change the initial learning rate to $10^{-6}$ and it decays to its $65\%$ for each 3000 steps. Figure \ref{fig:backward-semi learning dynamics} (a), all the state values decrease, and in (b) the loss is increasing in the later stage. As shown in Figure \ref{fig:backward-semi learning dynamics} (c), the highest and the lowest final state values  are traps (the terminal states with negative reward, black circle), and goal (the terminal state with a positive state, blue rectangle), respectively. As in the above analysis, a low estimation of terminal state  in Grid World or a high estimation of terminal state in Cart Pole indicates that state values are in a bad policy regime. Moreover, TD does not use the gradient at the terminal state to update the parameter, therefore, the backward-semi gradient dominates the value change in terminal states. 

The backward-semi gradient is the difference between TD with RG. The experiment in  Figure \ref{fig:backward-semi learning dynamics} confirms that this backward-semi gradient is the key that RG often fails. 


\section{Conclusion}

In this paper, we are trying to compare the performance of the TD and RG methods in a Deep Q-learning setting and analyze the reason that makes RG perform badly. We show their performance in four different environments, TD outperforms RG in all environments. Then we introduce noise to parameters  to show that models learned by TD are more robust. We also analyze two examples that represent two types of problems and conclude that RG learns a bad policy because it finds the solution in a bad regime where the Bellman residual error is small while the policy is bad.

There is an important fact that emerges from this work: a low Bellman residual is not enough for judging the goodness of a given policy,  and the state value should also be considered. The theory behind the policy regime is that the optimal criteria for the MDP problem are different from the solution of the Bellman optimal equation. This is a crucial difference between reinforcement learning and supervised learning wherein supervised learning loss represents its performance. This also inspires us, if we want to talk about generalization in reinforcement learning, we need to consider more indicators other than Bellman residual error.

This work also raises a cautionary tale about stochastic gradient descent (SGD) training, which is RG in this work. The extensive success of deep learning has formed an empirical belief that SGD or its variants are powerful to train neural networks to obtain good generalization. However, this work points out that in deep Q-learning with neural network approximation, SGD may have some serious drawbacks. Our work on the analysis of the backward-semi gradient also serves as an empirical basis for further theoretical analysis. 

There are still some questions unsolved: why backward-semi gradient has such learning dynamics? what is the function of the negative terminal state in the Grid World example? What is the relation between reward function design and performance? Why does TD ensure the desired state value form? Does TD prevent the bad regime for all kinds of problems? What will the policy regime be when the step reward function is more complex? We will try to answer these questions in the following works.

@article{sutton1988learning,
  title={Learning to predict by the methods of temporal differences},
  author={Sutton, Richard S},
  journal={Machine learning},
  volume={3},
  number={1},
  pages={9--44},
  year={1988},
  publisher={Springer}
}

@incollection{baird1995residual,
  title={Residual algorithms: Reinforcement learning with function approximation},
  author={Baird, Leemon},
  booktitle={Machine Learning Proceedings 1995},
  pages={30--37},
  year={1995},
  publisher={Elsevier}
}

@article{mnih2013playing,
  title={Playing atari with deep reinforcement learning},
  author={Mnih, Volodymyr and Kavukcuoglu, Koray and Silver, David and Graves, Alex and Antonoglou, Ioannis and Wierstra, Daan and Riedmiller, Martin},
  journal={arXiv preprint arXiv:1312.5602},
  year={2013}
}

@inproceedings{schoknecht2003td,
  title={TD (0) converges provably faster than the residual gradient algorithm},
  author={Schoknecht, Ralf and Merke, Artur},
  booktitle={Proceedings of the 20th International Conference on Machine Learning (ICML-03)},
  pages={680--687},
  year={2003}
}

@inproceedings{li2008worst,
  title={A worst-case comparison between temporal difference and residual gradient with linear function approximation},
  author={Li, Lihong},
  booktitle={Proceedings of the 25th international conference on machine learning},
  pages={560--567},
  year={2008}
}

@inproceedings{saleh2019deterministic,
  title={Deterministic Bellman residual minimization},
  author={Saleh, Ehsan and Jiang, Nan},
  booktitle={Proceedings of Optimization Foundations for Reinforcement Learning Workshop at NeurIPS},
  year={2019}
}

@article{zhang2019deep,
  title={Deep residual reinforcement learning},
  author={Zhang, Shangtong and Boehmer, Wendelin and Whiteson, Shimon},
  journal={arXiv preprint arXiv:1905.01072},
  year={2019}
}

@article{lillicrap2015continuous,
  title={Continuous control with deep reinforcement learning},
  author={Lillicrap, Timothy P and Hunt, Jonathan J and Pritzel, Alexander and Heess, Nicolas and Erez, Tom and Tassa, Yuval and Silver, David and Wierstra, Daan},
  journal={arXiv preprint arXiv:1509.02971},
  year={2015}
}

@article{brockman2016openai,
  title={Openai gym},
  author={Brockman, Greg and Cheung, Vicki and Pettersson, Ludwig and Schneider, Jonas and Schulman, John and Tang, Jie and Zaremba, Wojciech},
  journal={arXiv preprint arXiv:1606.01540},
  year={2016}
}

@article{silver2017mastering,
  title={Mastering the game of go without human knowledge},
  author={Silver, David and Schrittwieser, Julian and Simonyan, Karen and Antonoglou, Ioannis and Huang, Aja and Guez, Arthur and Hubert, Thomas and Baker, Lucas and Lai, Matthew and Bolton, Adrian and others},
  journal={nature},
  volume={550},
  number={7676},
  pages={354--359},
  year={2017},
  publisher={Nature Publishing Group}
}

@article{vinyals2019grandmaster,
  title={Grandmaster level in StarCraft II using multi-agent reinforcement learning},
  author={Vinyals, Oriol and Babuschkin, Igor and Czarnecki, Wojciech M and Mathieu, Micha{\"e}l and Dudzik, Andrew and Chung, Junyoung and Choi, David H and Powell, Richard and Ewalds, Timo and Georgiev, Petko and others},
  journal={Nature},
  volume={575},
  number={7782},
  pages={350--354},
  year={2019},
  publisher={Nature Publishing Group}
}

@inproceedings{deng2021unified,
  title={Unified conversational recommendation policy learning via graph-based reinforcement learning},
  author={Deng, Yang and Li, Yaliang and Sun, Fei and Ding, Bolin and Lam, Wai},
  booktitle={Proceedings of the 44th International ACM SIGIR Conference on Research and Development in Information Retrieval},
  pages={1431--1441},
  year={2021}
}

@article{bello2016neural,
  title={Neural combinatorial optimization with reinforcement learning},
  author={Bello, Irwan and Pham, Hieu and Le, Quoc V and Norouzi, Mohammad and Bengio, Samy},
  journal={arXiv preprint arXiv:1611.09940},
  year={2016}
}

@article{khalil2017learning,
  title={Learning combinatorial optimization algorithms over graphs},
  author={Khalil, Elias and Dai, Hanjun and Zhang, Yuyu and Dilkina, Bistra and Song, Le},
  journal={Advances in neural information processing systems},
  volume={30},
  year={2017}
}

@inproceedings{van2016deep,
  title={Deep reinforcement learning with double q-learning},
  author={Van Hasselt, Hado and Guez, Arthur and Silver, David},
  booktitle={Proceedings of the AAAI conference on artificial intelligence},
  volume={30},
  number={1},
  year={2016}
}

@inproceedings{wang2016dueling,
  title={Dueling network architectures for deep reinforcement learning},
  author={Wang, Ziyu and Schaul, Tom and Hessel, Matteo and Hasselt, Hado and Lanctot, Marc and Freitas, Nando},
  booktitle={International conference on machine learning},
  pages={1995--2003},
  year={2016},
  organization={PMLR}
}

@inproceedings{hessel2018rainbow,
  title={Rainbow: Combining improvements in deep reinforcement learning},
  author={Hessel, Matteo and Modayil, Joseph and Van Hasselt, Hado and Schaul, Tom and Ostrovski, Georg and Dabney, Will and Horgan, Dan and Piot, Bilal and Azar, Mohammad and Silver, David},
  booktitle={Thirty-second AAAI conference on artificial intelligence},
  year={2018}
}

@inproceedings{gu2016continuous,
  title={Continuous deep q-learning with model-based acceleration},
  author={Gu, Shixiang and Lillicrap, Timothy and Sutskever, Ilya and Levine, Sergey},
  booktitle={International conference on machine learning},
  pages={2829--2838},
  year={2016},
  organization={PMLR}
}

@inproceedings{duan2021risk,
  title={Risk bounds and rademacher complexity in batch reinforcement learning},
  author={Duan, Yaqi and Jin, Chi and Li, Zhiyuan},
  booktitle={International Conference on Machine Learning},
  pages={2892--2902},
  year={2021},
  organization={PMLR}
}

@article{scherrer2010should,
  title={Should one compute the temporal difference fix point or minimize the bellman residual? the unified oblique projection view},
  author={Scherrer, Bruno},
  journal={arXiv preprint arXiv:1011.4362},
  year={2010}
}
\bibliographystyle{plainnat} 
\bibliography{references}  

\end{document}